\documentclass[a4paper, 10pt, conference]{ieeeconf}      % Use this line for a4
                \usepackage{caption}      
                \usepackage{float}% paper
\usepackage{url}
\usepackage{graphicx}
\usepackage{multirow}
\usepackage{siunitx}
\usepackage{algorithmic}
\usepackage[colorlinks,linkcolor=blue]{hyperref}
\usepackage{color}
\definecolor{Red}{rgb}{1,0,0}
\definecolor{Blue}{rgb}{0,0,1}
\newcommand{\TRed}[1]{\textcolor{Red}{#1}}
\newcommand{\TBlue}[1]{\textcolor{Blue}{#1}}
\newcommand{\TBW}{28pt}
\usepackage{CJK}
\IEEEoverridecommandlockouts                              % This command is only
\title{\LARGE \bf
Active Terahertz Imaging Dataset for Concealed Object Detection
}

\author{\parbox{16cm}{\centering
    {\large Dong Liang, Fei Xue, and Ling Li}\\
    {\normalsize
    College of Computer Science and Technology, Nanjing University of Aeronautics and Astronautics\\}}
    \thanks{liangdong@nuaa.edu.cn}% <-this % stops a space
}

\begin{document}

\pagestyle{plain}
\iffalse
\begin{figure*}[ht]
	\includegraphics[width=1\textwidth]{image_wall_2.pdf}
	\caption{Visualization of the Terahertz object detection dataset. 11 classes of objects are labelled as shown in Table \ref{tab:det_class}.}
	\label{fig:det_class}       % Give a unique label
\end{figure*}
\maketitle
\fi

\twocolumn[{%
\renewcommand\twocolumn[1][]{#1}%
\maketitle
\begin{center}
    \centering
    \captionsetup{type=figure}
    \includegraphics[width=1\textwidth]{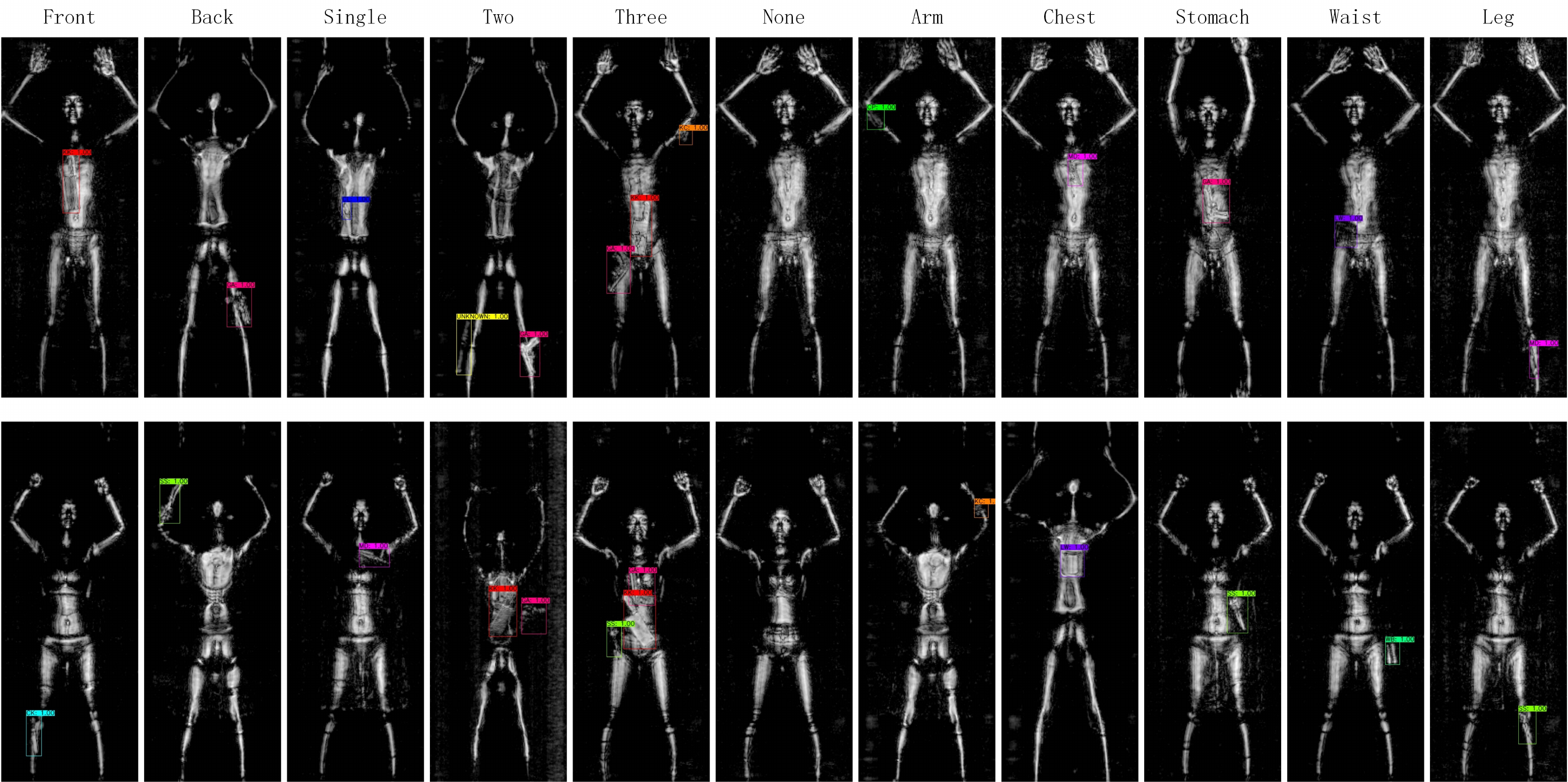}
    \captionof{figure}{Visualization of diversified samples in the dataset. From left to right, the front and back view, from 0 - 3 objects,  object on arm, chest, stomach, waist and leg.}
    	\label{fig:det_type}  
\end{center}%
}]
%%%%%%%%%%%%%%%%%%%%%%%%%%%%%%%%%%%%%%%%%%%%%%%%%%%%%%%%%%%%%%%%%%%%%%%%%%%%%%%%
\begin{abstract}

Concealed object detection in Terahertz imaging is an urgent need for public security and counter-terrorism. In this paper, we provide a public dataset for evaluating multi-object detection algorithms in active Terahertz imaging resolution 5 mm by 5 mm.  To the best of our knowledge, this is the first public Terahertz imaging dataset prepared to evaluate object detection algorithms. Object detection on this dataset is much more difficult than on those standard public object detection datasets due to its inferior imaging quality. Facing the problem of imbalanced samples in object detection and hard training samples, we evaluate four popular detectors: YOLOv3, YOLOv4, FRCN-OHEM, and RetinaNet on this dataset. Experimental results indicate that the RetinaNet achieves the highest mAP. In addition, we demonstrate that hiding objects in different parts of the human body affect detection accuracy. The dataset is available at https://github.com/LingLIx/THz\_Dataset.
\end{abstract}
\begin{figure*}[ht]
	\includegraphics[width=1\textwidth]{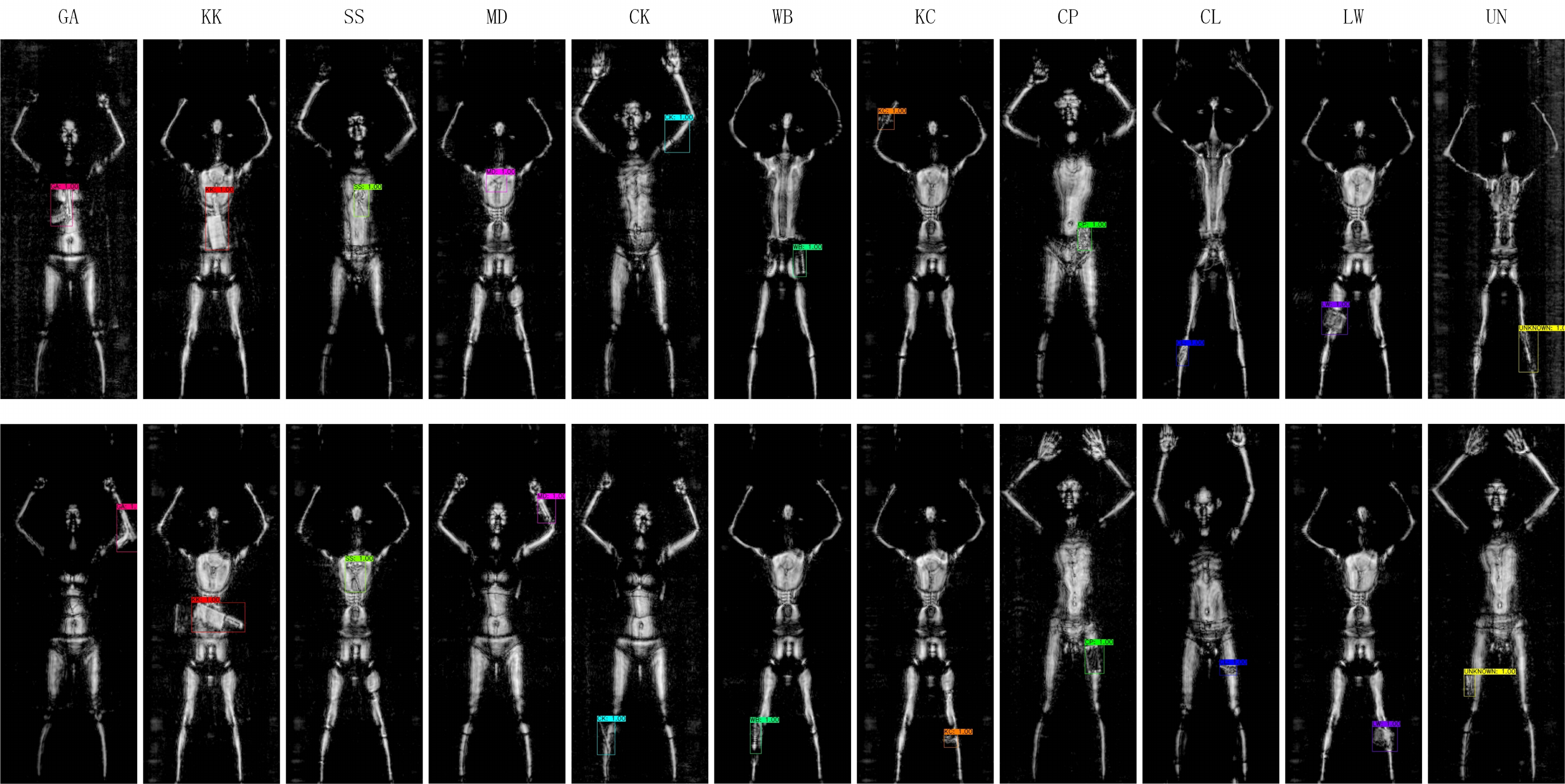}
	\caption{The case of image with only one object. 11 classes of objects are labelled as shown in Table \ref{tab:det_class}.}
   	\label{fig:det_class}   % Give a unique label
\end{figure*}
\begin{table*}[htb]
	\caption{Object classes and quantity of the dataset.}
	\label{tab:det_class}       % Give a unique label
	\scalebox{.99}{
		\centering
		\begin{tabular}{p{30pt}|p{10pt}p{\TBW}p{\TBW}p{30pt}p{\TBW}p{\TBW}p{\TBW}p{\TBW}p{\TBW}p{\TBW}p{\TBW}p{\TBW}}
			\hline\noalign{\smallskip}
			
			Class&GA&KK&SS&MD&CK&WB&KC&CP&CL&LW&UN& In total \\
			
			\noalign{\smallskip}\hline\noalign{\smallskip}
			
			Item&Gun&Kitchen Knife&Scissors&Metal Dagger&Ceramic Knife&Water Bottle&Key Chain&Cell Phone&Cigarette Lighter&Leather Wallet&Unknown&-- \\
			\hline\noalign{\smallskip}
			Quantity&116&100&96&64&129&107&78&129&163&78&289&1349 \\
			\noalign{\smallskip}\hline
	\end{tabular}}
\end{table*}
\begin{table*}[!ht]
	\caption{Details and statistics of the dataset.}
	\label{tab:det_static}       % Give a unique label
	\scalebox{.94}{
		\centering
		\begin{tabular}{lllllllllll}
			%{p{70pt}p{50pt}p{40pt}p{35pt}p{30pt}p{45pt}p{45pt}p{45pt}p{43pt}} 
			\hline\noalign{\smallskip}
			Image quantity & Image size &Imaging&Model&Object&Objects&Maximum&Minimum&Average&Labeling&Training $\&$\\
			&$\&$ format& resolution&gender&classes&per image&objects size&objects size&objects size&format& testing set\\
			\hline\noalign{\smallskip}
			3157 in total&&&&&&&&&&\\
			1218 with objects& 335$\times$880 p.x. & 5$\times$5 mm & 4 males &11&0,1,2,3& 13390 p.x. & 390 p.x. & 3222 p.x.&VOC&2555 training\\
			1349 without& JPEG&&6 females&&&&&&&602 testing\\
			\noalign{\smallskip}\hline
	\end{tabular}}
\end{table*}
\begin{figure}[!ht]
	\centering
	\includegraphics[width=0.49\textwidth]{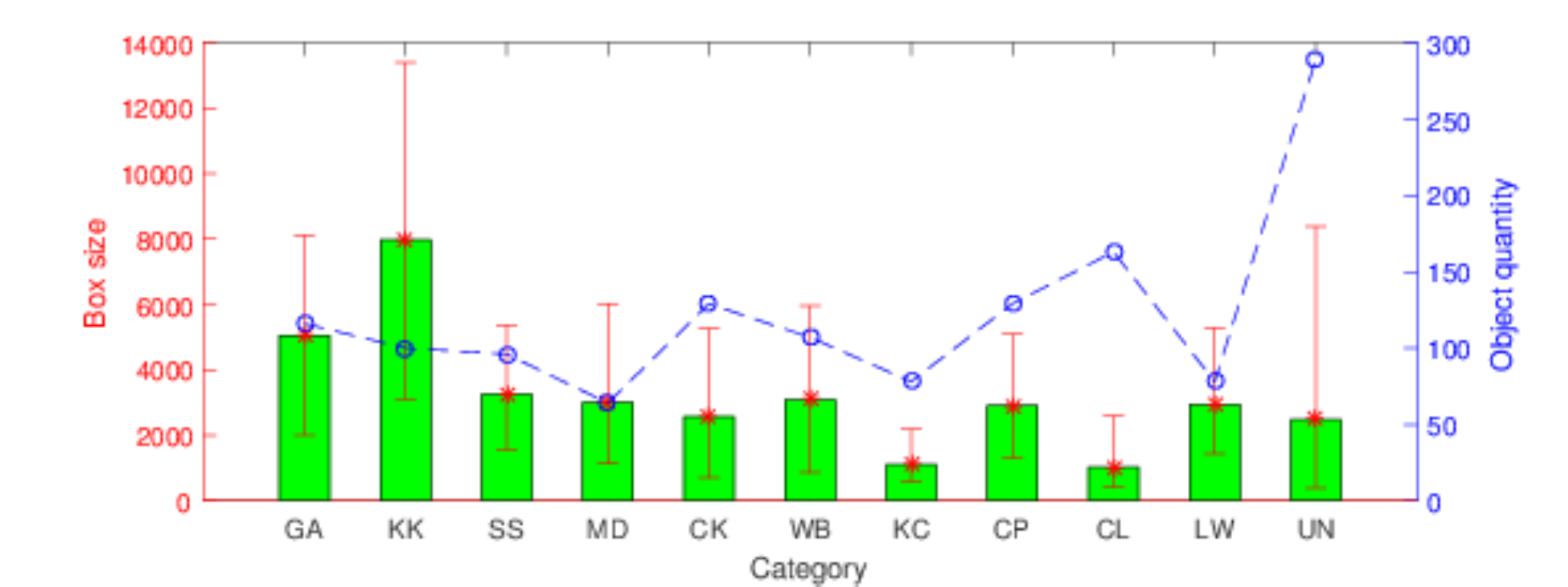}
	\caption{Size and quantity of different class in the dataset.}
	\label{fig:det_static}       % Give a unique label
\end{figure}

%%%%%%%%%%%%%%%%%%%%%%%%%%%%%%%%%%%%%%%%%%%%%%%%%%%%%%%%%%%%%%%%%%%%%%%%%%%%%%%%
\section{INTRODUCTION}

Detecting concealed objects underneath clothing is an essential step in public security checks, while the manual check is criticized for inefficiency, invasion of privacy, and high rate of missed detection. Terahertz band, between microwave and infrared, with the frequency range from 0.1 to 10 Terahertz \cite{helal2021signal}, provides a non-contact way to discover objects concealed underneath clothing with no harm to human health. 
According to the presence or absence of Terahertz source irradiation, there are two categories of Terahertz imaging systems – passive \cite{kowalski2015passive,yeom2011real,shen2008detection} and  active \cite{cooper2009approach,yan2015broadband,liang2019concealed}. 
Although the active terahertz imaging technology has advantages in anti-interference and signals stability over the passive terahertz imaging technology, the imaging contrast and signal-to-noise ratio are still technical bottlenecks. Although there are several methods \cite{yeom2011real,kowalski2015passive,liang2019concealed, cooper2009approach, shen2008detection} proposed for object detection for active/passive Terahertz imaging, all of them utilized private dataset to evaluate the performance. To our knowledge, there is no public dataset in Terahertz imaging to evaluate object detection algorithm. In this paper, we provide an active Terahertz imaging dataset for multi-object detection with 3157 image samples with 1347 concealed objects. We hope it can offer opportunities to link the computer vision and photoelectric imaging community to reveal the shortcomings of the existing object detection methods, and further develop strategies more suitable for harsh imaging conditions. In the following parts, we first describe the characteristics of the dataset, and then describe the four detectors used.The performance of each detector in this data set is revealed in experiments. We also have a preliminary experiments and discussion on the image-level evaluation, the position and posture of the objects.

%%%%%%%%%%%%%%%%%%%%%%%%%%%%%%%%%%%%%%%%%%%%%%%%%%%%%%%%%%%%%%%%%%%%%%%%%%%%%%%%
\section{THE DATASET}

An active Terahertz imaging system developed by China Academy of Engineering Physics is used for image acquisition. This imaging system adopts array scanning mode, works at 140 GHz, with imaging resolution 5 mm by 5 mm. When acquiring data, human models stand with hidden objects in their clothing. This dataset is diversified – the number of hidden objects in an image is from 0 to 3. There are 4 male and 6 female models with an equal amount of participation during image acquisition. Objects are hidden in different positions of the human body. Images acquired in each imaging include the front and back of the model. Fig \ref{fig:det_class} shows some visual annotations of each category. 11 classes of objects and their corresponding quantity of objects are labeled as shown in Table \ref{tab:det_class}. Note that the Class Unknow (UN) refers to those objects that do not fall into the 10 clear classes. We annotated the bounding boxes, and class labels of the dataset in Pascal VOC format \cite{everingham2010pascal}. Fig \ref{fig:det_type} shows some visual annotations of diversification. The statistical result of the Terahertz dataset is shown in Table \ref{tab:det_static}. The statistical result of the object size and quantity distribution in the dataset is shown in Fig \ref{fig:det_static}. When we collect data, we pay more attention to various situations that may occur in actual applications so that the test results are more in line with the real scene. For example, we consider the model's figure, the thickness and material of the clothes, the size, quantity, material and placement of the objects, etc.

\section{THE DETECTORS}
Object detection aims to infer the location, size, and class label of the object on an image. In this section, we discuss the detectors we use for the evaluation of this dataset. In this terahertz dataset, the problem of imbalance training samples (mainly because of the detector' structure) and hard training samples (mainly because of the ambiguous appearance), we evaluate four state-of-the-art detectors: YOLOv3, YOLOv4, FRCN-OHEM, and RetinaNet on this dataset.

\subsection{YOLOv3 and YOLOv4}

YOLOv3 \cite{redmon2018yolov3} is the third edition of You Only Look Once (YOLO) series detector, which is widely used in the industrial field. It uses a 52 layers feature extraction network with darknet-53 backbone \cite{redmon2017yolo9000} with residual jump connections. To avoid the negative effect of gradient caused by pooling, it abandons the pooling operation and uses convolution to achieve down-sampling. It uses a feature pyramid network \cite{lin2017feature} to create multiple scales feature maps. To speed up the computation, each branch detects in the corresponding grid cell, and each grid cell predicts 3 boxes and the probability of each class. 

YOLOv4 \cite{bochkovskiy2020yolov4} uses CSPDarknet53 \cite{wang2020cspnet} as its backbone, Spatial Pyramid Pooling \cite{he2015spatial, chen2017deeplab} as its pooling module, PANet \cite{liu2018path} as the path-aggregation neck. For training, DropBlock \cite{ghiasi2018dropblock} is selected as regularization method. In addition, YOLOv4 uses self-adversarial training for data augmentation and verified that CIoU \cite{zheng2020distance} can achieve better convergence speed and accuracy on the bounding box regression problem. YOLOv4 shows better detection accuracy and speed than YOLOv3.

\subsection{FRCN-OHEM}

Online Hard Example Mining (OHEM) \cite{shrivastava2016training} is designed for selecting hard samples for training. It is based on Fast R-CNN, a two-stage object detector that first proposes candidate regions and then classifies them. FRCN-OHEM has two similar regions of interested (RoI) networks. Network (a) is read-only and only does forward calculations for all RoIs. Network (b) does both read and write and forward and reverse propagation calculations for the selected hard RoIs. In a training iteration, first, the feature maps are calculated, the network (a) performs forward calculation on all RoIs and calculates the loss, then hard RoIs are selected and feed into the readable and writable network (b) to perform forward calculation and reverse propagation to update the network. Finally, the network parameters are assigned to the readable network. In summary, OHEM focuses on the training of hard samples by dual RoI networks interaction.

\subsection{RetinaNet}
RetinaNet \cite{lin2017focal} is a one-stage object detector, in which the Resnet\cite{he2016deep} is selected as the backbone. A feature pyramid is adopted to solve the multi-scale detection. Generally, the one-stage detector is faster but less robust than the two-stage detector. The imbalance of training samples mainly causes the performance difference between them. In the two-stage detectors, many negative samples are filtered out through the fixed proportion of positive and negative samples in the classification and regression stage, or the positive and negative are relatively balanced through OHEM. While in a one-stage detector, the training is dominated by a large number of negative samples which meets the typical class imbalance problem. When training to minimize the loss function, the introduction of many negative samples often causes bias, resulting in the decline of detection accuracy. In RetinaNet, Focal Loss (FL) is designed to solve the imbalance problem with a sample balance weight; meanwhile, it also considers the hard sample case with a hard sample weight. To a certain extent, this designation alleviates the class imbalance problem and the hard sample training problem.

\begin{table*}[ht]
	\caption{Quantitative results of detection experiments. The best result is in red whereas the worst one is in blue.}
	\label{tab:3}       % Give a unique label
	
	\centering
	\scalebox{.72}{
		\begin{tabular}{lllllllllllllllllll}
			\hline\noalign{\smallskip}
			
			Index&Model&Backbone&Loss&Speed (FPS)&Code framework&NMS&GA&KK&SS&MD&CK&WB&KC&CP&CL&LW&UN&mAP(\%) \\
			
			\noalign{\smallskip}\hline\noalign{\smallskip}
			
			1&YOLOv3   		&Darknet-53		&CE		&3.85 			&Tensorflow2.0 	&CPU  &67.15			&\TBlue{81.69}	&45.56			&\TBlue{0.0}	&16.39			&\TBlue{11.29}	&25.00			&\TBlue{60.74}	&\TBlue{18.57}	&\TRed{75.43}	&30.35 			&\TBlue{39.29}	\\
			2&YOLOv4		&CSPDarknet53	&CE				&10.32			&PyTorch		&CPU&\TBlue{58.05}	&82.72			&\TRed{67.27}	&\TBlue{0.0}	&\TBlue{12.50}	&34.72			&\TBlue{8.33}			&61.62			&54.79			&61.02			&\TBlue{14.23} 	&41.39	\\
			3&FRCN-OHEM     &VGG16     		&CE				&\TBlue{2.50}	&Keras		   	&CPU  &\TRed{83.39}	&\TBlue{81.69}	&66.67			&\TBlue{0.0}	&16.39			&31.56			&37.38				&63.21			&\TBlue{18.57}	&44.23			&22.38 			&42.32	\\
			4&RetinaNet		&ResNet-50 		&CE				&\TRed{15.05}	&Keras			&GPU &\TRed{83.39}	&\TRed{100.0}	&45.00			&\TBlue{0.0}	&\TRed{48.73}	&57.14			&83.33					&63.21			&42.28			&44.23			&33.02 			&54.58	\\
			5&RetinaNet		&ResNet-101		&CE				&10.19			&Keras			&GPU &73.33			&88.89			&\TBlue{39.50}	&\TBlue{0.0}	&21.08			&47.22			&80.56					&67.62			&47.61			&\TBlue{40.00}	&22.38 			&47.99	\\
			6&RetinaNet		&ResNet-50 		&FL				&\TRed{15.05}	&Keras			&GPU &82.05			&88.56			&66.67			&\TBlue{0.0}	&37.16			&\TRed{64.86}	&78.57					&68.75			&62.87			&65.17			&\TRed{34.34} 	&59.00\\
			7&RetinaNet		&ResNet-101		&FL			&11.70			&Keras			&GPU &72.31			&88.89			&44.00			&\TBlue{0.0}	&45.57			&57.78			&53.70			&\TRed{83.33}	&56.51			&62.87			&24.67			&53.60	\\
			8&RetinaNet+P2	&ResNet-50		&FL		&14.90			&Keras			&GPU &82.05			&88.56			&66.67			&\TBlue{0.0}	&37.16			&\TRed{64.86}	&\TRed{85.05}			&68.75			&\TRed{70.87}	&65.17			&\TRed{34.34} 	&\TRed{60.32}	\\
			
			\noalign{\smallskip}\hline
	\end{tabular}}
\end{table*}

%%%%%%%%%%%%%%%%%%%%%%%%%%%%%%%%%%%%%%%%%%%%%%%%%%%%%%%%%%%%%%%%%%%%%%%%%%%%%%%%
\section{EXPERIMENTS}
\subsection{Settings}
YOLOv3, YOLOv4, FRCN-OHEM, RetinaNet  models are compared. RetinaNet is also compared using different loss functions and backbones. In RetinaNet, we also add a lower-level feature layer (P2) to deal with small object detection since it preserves more detailed feature information. What can be expected is that the embedding of a feature layer will increase computing cost when inferencing, which is also confirmed in the subsequent experiments. Mean Average Precision (mAP) and Average Precision (AP) are used as the evaluation criteria (Pascal VOC's mAP after 2010). Specifically, the AP of each category is equal to the area under its Precision-Recall curve, and the mAP is the mean AP of multiple categories. We used the original pre-trained model for all object detection models that participated in the experiment and fine-tuned it on our training set. The model parameters are set strictly following the default parameters.
\subsection{Comparison study}
\begin{figure}[ht]
	\centering
	\includegraphics[width=1\linewidth]{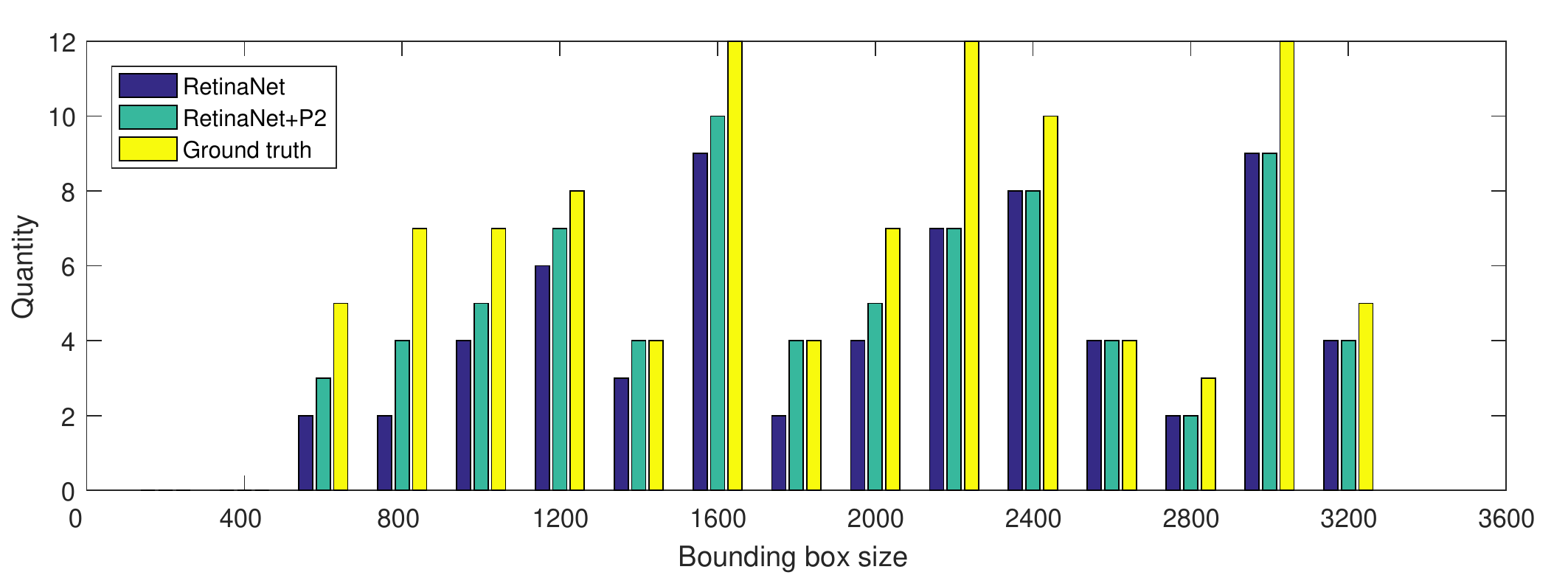}
	\caption{True positive sample statistics on test result. Only the area of the bounding box below 3600 is shown.}
	\label{fig:4}       % Give a unique label
\end{figure}

As it is shown in Table \ref{tab:3}, the mAP of RetinaNet is higher than that of YOLOv3 and YOLOv4 when Cross-Entropy (CE) is used. We can notice that the detection accuracy of RetinaNet using Focal Loss (FL) is higher than Cross-Entropy. Because the code of YOLOs we use is duplicated and it uses CPU for the NMS process, the detection speed of YOLOs is lower than RetinaNet. The AP of FRCN-OHEM is higher than YOLOs but lower than RetinaNet. The detection speed of FRCN-OHEM is slower than Retinanet. In Table \ref{tab:3}, it is noted that RetinaNet with the backbone as ResNet-101 does not achieve AP as high as RetinaNet with the backbone as ResNet-50. The causal of this phenomenon may be the quantity of the training samples. In the dataset, although there are 2555 images for training, the samples of each class are limited. In the case of limited training data, the introduction of a deeper network would not improve the performance but would lead to model over-fitting. For specific classes, RetinaNet has better AP than YOLOs in many categories, especially for the CK (Ceramic Knife) and KC (Key Chain), which are difficult to distinguish. The mAP of all models for MD (Metal Dagger) is 0; that may because the number of MD in the dataset is small, and its appearance is very close to CK (Ceramic Knife). The experimental results in the last row show that the added feature layer (P2) can effectively improve the detection accuracy of small objects, as shown in Fig \ref{fig:4}.

\subsection{Image-level detection}

In practice, the detection rate and false alarm rate are two significant indicators to measure the system's performance. Therefore, we define object-level detection indicators for one image as follows: 

\textbf{ObjTurePositive:} the detector marks the location of the hidden object, and the IoU between the detection bounding box and the ground truth bounding box is more than 50\%. 

\textbf{ObjFalsePositive:} the detector marks a location of the hidden object, but there is no object. Then we define image-level detection indicators as follows: 

\textbf{ImgDetection:} if some or all of the hidden objects in a Terahertz image are ObjDetection, it is ImgDetection. 

\textbf{ImgFalseAlarm:} if there is any ObjFalseAlarm in a Terahertz image, it is ImgFalseAlarm. 

Finally, we get image-level Detection Rate (DR) and False Alarm rate (FA) as follows:
\begin{equation}
\label{equ:5}
DR=\frac{1}{n}\sum\limits_{i=1}^{n}{ImgDetection(i)}
\end{equation}
\begin{equation}
\label{equ:6}
FA=\frac{1}{n}\sum\limits_{i=1}^{n}{ImgFalseAlarm(i)}
\end{equation}
where $i$ is the image index and $n$ is the total quantity of images.

\begin{table}[ht]
	\caption{Image level detection result.  The best result is in red whereas the worst one is in blue.}
	\label{tab:image_level}       % Give a unique label
	\centering
	\scalebox{.95}{
	\begin{tabular}{lll}
		\hline\noalign{\smallskip}
		
		Model&Detection Rate (\%)&False Alarm rate (\%) \\
		
		\noalign{\smallskip}\hline\noalign{\smallskip}
		
		YOLOv4		&84.49				&\TRed{0.63}	\\
		FRCN-OHEM	&\TBlue{83.86}		&\TBlue{18.99}	\\
		RetinaNet	&\TRed{91.46}		&1.27			\\
		
		\noalign{\smallskip}\hline
	\end{tabular}}
\end{table}

The image-level detection result is shown in Table \ref{tab:image_level}. The detection rate (DR) of RetinaNet is over 90\% when its False Alarm rate (FA) is 1.27\%. Although YOLOv4 has the lowest FA, its DR is lower than RetinaNet. FRCN-OHEM has the worst performance. 

\subsection{Position analysis of object in Terahertz imaging}

Objects may be placed in different places on the human body. Meanwhile, the hidden objects may be placed perpendicular to the imaging plane rather than paralleled to the imaging plane. As shown in Table \ref{tab:pos}, we calculate the Recall (ture positive rate) in different positions of the human body. The last two columns indicate the imaging direction of the object. Table \ref{tab:pos} shows that the Recall rate of objects on arms is relatively low. Because of the imaging diversity of the human body background, the Recall is also low when objects are on the body. All detectors have better Recall of the objects parallel to the imaging plane than perpendicular to the imaging plane. Therefore, it is necessary to use multi-view detection to effectively avoid missing detection of objects placed perpendicular to the imaging plane.

\begin{table}[ht]
	\caption{Recall in different cases of position and direction. 147 objects in the test set are counted. The best result is in red whereas the worst one is in blue.} 
	\label{tab:pos}       % Give a unique label
	\centering
		\scalebox{.85}{
	\begin{tabular}{l|lll|ll}
		\hline\noalign{\smallskip}
		
		Model	
		&On Arms 	&On Body 	&On Legs 	&Parallel 	&Perpendicular \\
		
		\noalign{\smallskip}\hline\noalign{\smallskip}
			
		YOLOv4	
		&\TBlue{0.2500}	&0.4783			&0.6143			&0.5789			&0.4423 \\
		
		FRCN-OHEM
		&0.3750			&\TBlue{0.4638}	&\TBlue{0.5000}	&\TBlue{0.5158}	&\TBlue{0.4038}	\\
		
		RetinaNet		
		&\TRed{0.5000}	&\TRed{0.4928}	&\TRed{0.7000}	&\TRed{0.6105}	&\TRed{0.5577}	\\
		
		\noalign{\smallskip}\hline
	\end{tabular}}
\end{table}

%%%%%%%%%%%%%%%%%%%%%%%%%%%%%%%%%%%%%%%%%%%%%%%%%%%%%%%%%%%%%%%%%%%%%%%%%%%%%%%%
\section{CONCLUSIONS}

In this paper, an active Terahertz imaging dataset for multi-object detection is provided. Experimental results indicate that the RetinaNet achieves the highest mAP. In addition, we demonstrate that hiding objects in different parts of the human body affect detection accuracy. We hope it can offer opportunities to link the computer vision and photoelectric imaging community. Future work will focus on exploring the detection methods of small objects on this active Terahertz images dataset.

%%%%%%%%%%%%%%%%%%%%%%%%%%%%%%%%%%%%%%%%%%%%%%%%%%%%%%%%%%%%%%%%%%%%%%%%%%%%%%%%

{\small
\bibliographystyle{plain}
\bibliography{egbib}
}

\end{document}